\documentclass[10pt,journal,compsoc]{IEEEtran}
\makeatletter
\def\endthebibliography{%
	\def\@noitemerr{\@latex@warning{Empty `thebibliography' environment}}%
	\endlist
}
\makeatother

\ifCLASSOPTIONcompsoc
  \usepackage[nocompress]{cite}
\else
  \usepackage{cite}
\fi

\usepackage{url}
\usepackage{epsfig}
\usepackage{graphicx}
\usepackage{subfigure}
\usepackage{amsmath}
\usepackage{amssymb}
\usepackage{amsfonts}
\usepackage{mathrsfs}
\usepackage{booktabs}
\usepackage{multirow, eucal}
\usepackage{helvet}
\usepackage{courier}
\usepackage{bm}
\usepackage{tabu}
\usepackage{bbm}
\usepackage{color}
\usepackage{epstopdf}
\usepackage{wrapfig}
\usepackage{picinpar}

\usepackage[vlined,ruled,linesnumbered]{algorithm2e}
\usepackage{cite}
\usepackage[pagebackref=true,breaklinks=true,letterpaper=true,colorlinks,bookmarks=false]{hyperref}

\usepackage{caption}

\def\etal{\emph{et al}\onedot}

\def\etal{{\em et al.\/}\, }

\def\0{{\mathbf 0}}
\def\1{{\mathbf 1}}

\def\bB{{\mathbf B}}

\def\bW{{\mathbf W}}

\def\bb{{\mathbf b}}

\def\bp{{\mathbf p}}

\def\bx{{\mathbf x}}
\def\by{{\mathbf y}}

\definecolor{mypink}{cmyk}{0, 0.7808, 0.4429, 0.1412}

\def\cL{{\cal L}}

\def\bWl{\bW_{\rm{low}}}
\def\bWf{\bW_{\rm{full}}}
\def\bpl{\bp_{\rm{low}}}
\def\bpf{\bp_{\rm{full}}}
\def\Ml{M_{\rm{low}}}
\def\Mf{M_{\rm{full}}}

\hypersetup{%
	colorlinks=true,
	linkcolor=blue,
	anchorcolor=blue,
	citecolor=blue}

\newcommand{\tabincell}[2]{\begin{tabular}{@{}#1@{}}#2\end{tabular}}

\begin{document}

\title{Effective Training of  Convolutional Neural Networks with Low-bitwidth Weights and  Activations
}

\author{Bohan Zhuang,
        Mingkui Tan,
	    Jing Liu,
        Lingqiao Liu,
        Ian Reid,
        and Chunhua Shen
\IEEEcompsocitemizethanks
{\IEEEcompsocthanksitem
B. Zhuang, J. Liu and C. Shen are  with Faculty of Information Technology, Monash University, Australia.
C. Shen is the corresponding author.
\protect\\
E-mail: \{firstname.lastname\}@monash.edu
\IEEEcompsocthanksitem
C. Shen, L. Liu and I. Reid are with  The University of Adelaide, Australia. \protect\\
E-mail: \{firstname.lastname\}@adelaide.edu.au
\IEEEcompsocthanksitem   M. Tan is  with South China University of Technology.\protect\\
E-mail:  mingkuitan@scut.edu.cn

\IEEEcompsocthanksitem
First three authors contributed equally to this work.
\IEEEcompsocthanksitem
\color{blue}
Appearing in \it IEEE Trans.\ Pattern Analysis and Machine Intelligence, 2021.
Content may be slightly different from the final published version.
}%
}

\IEEEtitleabstractindextext{%
\begin{abstract}
This paper tackles the problem of training a deep convolutional neural network of both
low-bitwidth weights and activations.
Optimizing a low-precision network is very challenging due to the non-differentiability of the quantizer, which may result in substantial accuracy loss. To address this, we propose three practical approaches, including (i) progressive quantization; (ii) stochastic precision;
and (iii) joint knowledge distillation to improve the network training.
First, for progressive quantization, we propose two schemes to progressively find good local minima.  Specifically, we propose to first optimize a network with quantized weights and subsequently quantize activations. This is in contrast to the traditional methods which optimize them simultaneously. Furthermore, we propose a second progressive quantization scheme which gradually decreases the bit-width from high-precision to low-precision during training.
Second, to alleviate the excessive training burden due to the multi-round training stages, we further propose a one-stage stochastic precision strategy to randomly sample and quantize sub-networks while keeping other parts in full-precision.
Finally, we adopt a novel learning scheme to jointly train a full-precision model alongside the low-precision one. By doing so, the full-precision model provides hints to guide the low-precision model training and significantly improves the performance of the low-precision network.
Extensive experiments on various datasets (\textit{e.g.}, CIFAR-100, ImageNet) show the effectiveness of the proposed methods.
\end{abstract}

}

\maketitle

\IEEEdisplaynontitleabstractindextext
\IEEEpeerreviewmaketitle

\section{Introduction}

State-of-the-art deep neural networks ~\cite{krizhevsky2012imagenet, simonyan2014very, he2016deep} usually involve millions of parameters and need billions of FLOPs %
for training and inference.
The significant memory consumption and computational cost can %
make it intractable to deploy models to
 mobile, embedded hardware devices.
 To improve computing and memory efficiency, various solutions have been proposed, including network pruning \cite{zhuang2018discrimination, he2017channel, li2017pruning}, low rank approximation of weights \cite{kim2015compression, zhang2016accelerating}, training a low-precision network \cite{zhou2017incremental, courbariaux2015binaryconnect, zhu2016trained, zhou2016dorefa} and efficient architecture design~\cite{guo2019single, howard2017mobilenets, zhang2017shufflenet}. In this work, we follow the idea of training a low-precision network and our focus is to improve the training process of such a network. 
Thus, our work targets the problem of training a network with both extremely low-bit weights and activations.

The solutions proposed in this paper contain three components. They can be applied independently or jointly. The first component is the progressive quantization which consists of two schemes. 
The first strategy is to adopt a two-stage training process. At the first stage, only the weights of a network are quantized. After obtaining a sufficiently good solution 
at 
the first stage, the activation of the network is further required to be in low-precision and the network 
is 
trained again. Essentially, this two-stage approach first solves a related sub-problem, i.e., training a network with only low-bit weights and the solution of the sub-problem provides a good initial point for training our target problem. Following the similar idea, we propose our second scheme by performing progressive training on the bitwidth aspect of the network. Specifically, we incrementally train a serial of networks with the quantization bitwidth (precision) gradually decreased from full-precision to the target precision.  

However, the above progressive quantization needs several retraining steps which introduces additional training burdens. To solve this problem, we further propose our second component termed 
stochastic precision to effectively combine these two strategies into 
one
single
training stage. Inspired by dropout strategies~\cite{srivastava2014dropout, huang2016deep}, we randomly select a portion of the model (%
\textit{e.g.}, layers, blocks) and activations or weights to quantize while keeping other parts in full-precision.
Thus, 
we can  improve the gradient flow for effectively training quantized neural networks.

The third component is inspired by the recent progress of  information distillation \cite{romero2014fitnets, hinton2015distilling, parisotto2016actor, zagoruyko2016paying, ba2014deep}. The basic idea of those works is to train a target network alongside another guidance network. For example, the works in \cite{romero2014fitnets, hinton2015distilling, parisotto2016actor, zagoruyko2016paying, ba2014deep} propose to train a small student network to mimic the deeper or wider teacher network. They add an additional regularizer by minimizing the difference between student's and teacher's posterior probabilities~\cite{hinton2015distilling} or intermediate feature representations~\cite{ba2014deep, romero2014fitnets}. It is observed that by using the guidance of the teacher model, better performance can be obtained with the student model than directly training the student model on the target problem.  Motivated by these observations, we propose to train a full-precision network alongside the target low-precision network.
In our work, the student network has the similar topology as that of the teacher network, except that the student network is low-precision while the teacher network keeps full-precision operations. Moreover, in contrast to standard knowledge distillation methods, we allow the teacher network to be jointly optimized with the student network rather than being fixed since we discover that this 
strategy 
enables the two networks adjust better to each other. Interestingly, the performance of both the full-precision teacher and the low-precision student can be improved. 

Our main contributions are 
summarized as follows.  
\begin{itemize}
    \item  We propose two progressive quantization schemes for tackling the non-differentiability of quantization operations during training. In the first scheme, we propose a two-stage training manner, where the weights are first quantized to serve as a good initialization on further quantizing activations. In the second scheme, we progressively reduce the bitwidth during training to find better local minima. 
    
    \item To reduce the extra training burden, we introduce structured stochastic training,
    leading to an effective, simplified one-stage training approach.

    \item To  our knowledge, we are the first to propose to improve the low-precision network training using knowledge distillation technique where the full-precision teacher and the quantized student are jointly optimized to adapt to each other. We explore different distilling schemes in Sec.~\ref{exp:experiment} and all produce improved accuracy for the low-precision model. 
    
    \item We conduct extensive experiments %
    with
    various precisions and architectures on the image classification task.

\end{itemize}

This paper 
extends the preliminary conference version~\cite{zhuang2018towards} in several aspects. 1) Although the multi-stage progressive quantization in~\cite{zhuang2018towards} clearly improves the performance,
the multiple re-initialization and fine-tuning steps 
make the training procedure complex and introduce computation overhead. 
To solve this problem, here we propose a much simpler one-stage stochastic precision strategy that enjoys the advantage of the multi-stage progressive quantization.
2) We extend the hint-based joint knowledge distillation to a more advanced framework that unifies attention transfer \cite{zagoruyko2016paying} and posterior-based schemes. 
3) We now conduct extensive experiments on ImageNet over various architectures to formulate strong and comprehensive baselines for future works. We study several schemes which produce low-precision networks using different distilling strategies and provide interesting analysis.

\section{Related work}
\label{sec:related_work}

We have witnessed a growing interest of model compression methods, such as limited numerical precision, efficient architecture design and knowledge distillation.
We also study the dropout strategies in this paper.
Next we discuss related literature with respect to these aspects.

\noindent\textbf{Limited numerical precision.}
Model quantization aims to quantize the weights, activations and even backpropagation gradients into low-precision, to yield highly compact DNNs compared to their floating-point counterparts. As a result, most of the multiplication operations in network inference can be replaced by more efficient addition or bitwise operations. In general, quantization methods generally involve binary neural networks (BNNs) and fixed-point quantization. 
In particular, BNNs \cite{rastegari2016xnor, hubara2016binarized, bethge2018training,bethge2018learning,tang2017train,guo2017network,Liu_2018_ECCV}, where both weights and activations are quantized to binary tensors, are reported to have potentially $32\times$ memory compression ratio, and up to $58\times$ speed-up on CPU compared with the full-precision counterparts. However, BNNs still suffer from sizable performance drop issue, hindering them from being widely deployed.   
To make a trade-off between accuracy and complexity, researchers also study fixed-point quantization \cite{zhou2016dorefa, choi2018pact,esser2019learned,zhuang2019structured, lin2017towards}. In general, quantization algorithms aim at tackling two core challenges. The first challenge is to design accurate quantizers to minimize the information loss. Early works use handcrafted heuristic quantizers \cite{zhou2016dorefa} while later studies propose to adjust the quantizers to the data, basically based on matching the original data distribution \cite{zhou2016dorefa, Cai_2017_CVPR}, minimizing the quantization error \cite{zhang2018lq} or directly optimizing the quantizer with stochastic gradient descent \cite{choi2018pact, jung2019learning}. 
The second challenge is to approximate gradients of the non-differentiable quantizer.
To solve this problem, most works in literature simply employ ``pseudo-gradients'' according to the straight-through estimator (\textit{i.e.}, STE)~\cite{bengio2013estimating}. Some 
recent
studies propose to improve the discrete optimization problem via loss-aware training \cite{hou2018loss}, regularization \cite{ding2019regularizing, Yoojin2018regular, bai2019proxquant}, entropy maximization \cite{park2017weighted, polino2018model}, or smoothing the quantizer \cite{louizos2019relaxed}.
In addition to the quantization algorithms design, the underlying implementation and acceleration libraries \cite{ignatov2018ai,yang2017bmxnet,umuroglu2017finn,jacob2017quantization} are indispensable to expedite the quantization technique to be deployed on energy-efficient edge devices.
In this paper, we propose three training solutions that can be built upon general quantization approaches.

\noindent\textbf{Efficient architecture design.}
The increasing demand for %
highly energy efficient neural networks that are deployable to embedded hardware devices has motivated the network architecture design.
SqueezeNet~\cite{iandola2016squeezenet} 
replaces $3\times3$ convolutional filters with $1\times1$ size, which 
significantly 
decreases  the complexity. %
Depthwise separable convolutions employed in Xception~\cite{chollet2017xception},  MobileNet~\cite{howard2017mobilenets} and ShuffleNet~\cite{zhang2017shufflenet} have been proved to be efficient and effective. 
Since it is infeasible to manually explore the optimal architecture from the enormous design space, neural architecture search (NAS) aims at automating the architecture design, giving rise to methods based on the reinforcement learning~\cite{zoph2016neural, pham2018efficient, zoph2017learning, liu2017progressive}, evolutionary algorithms \cite{real2019regularized}, or gradient-based methods \cite{liu2019darts, cai2019proxylessnas, guo2019single}.

Moreover, network pruning \cite{zhuang2018discrimination, yu2018nisp} 
can be viewed as a special case of 
NAS, aiming to remove redundant connections such as convolutional filters. 
Some works also employ reinforcement learning~\cite{ashok2018n2n, he2018amc}, Bayesian optimization \cite{tung2018clip} or NAS \cite{dong2019network} to automatically search the pruning policy for each layer.

\noindent\textbf{Knowledge distillation.}
Knowledge distillation was initially proposed for model compression, where a powerful wide/deep teacher distills knowledge into a narrow/shallow student to improve its performance~\cite{romero2014fitnets, hinton2015distilling}. In terms of the representation of knowledge to be distilled from the teacher, existing models typically use
teacher's class probabilities~\cite{hinton2015distilling} and/or feature representations~\cite{romero2014fitnets, zagoruyko2016paying}.  Knowledge distillation has been widely used in many computer vision tasks. Zhang~\etal~\cite{zhang2016real} proposes to transfer the knowledge learned with optical flow CNN to improve the action recognition performance. Moreover, several works propose to learn efficient object detection~\cite{chen2017learning, wei2018quantization} and semantic segmentation~\cite{he2019knowledge} with distillation. In contrast to previous approaches, we concentrate on improving the performance of the quantized neural network. By adapting the teacher and student altogether, we can steadily improve the performance of the quantized student network and even the full-precision teacher network.
Note that concurrent works ~\cite{mishra2018apprentice, polino2018model} with ours~\cite{zhuang2018towards} also apply knowledge distillation to quantization. We have also explored 
it together with other advanced training strategies.

\noindent\textbf{Dropout.}
Dropout~\cite{srivastava2014dropout}, Maxout~\cite{goodfellow2013maxout}, DropConnect~\cite{wan2013regularization} and DropIn~\cite{smith2016gradual} are a category of approaches that 
stochastically drop intermediate nodes or connections during training to prevent the network from overfitting. %
They essentially perform different types of regularization. 
Huang~\etal~\cite{huang2016deep} further propose stochastic depth regularization via randomly dropping a subset of layers during training. 
Dong~\etal~\cite{dong2017learning} proposes to randomly quantize a portion of weights to low-precision in the incremental training framework~\cite{zhou2017incremental}. The method in~\cite{dong2017learning} was developed for only quantizing weights of a network. In our method, we develop an extension of it by further randomly quantizing a portion of the network, \textit{i.e.},\textit{ layers or blocks as well as activations and weights}. Moreover, we \cite{zhuang2018towards} propose two progressive training strategies: 1) quantizing weights and activations in a two-stage manner; 2) progressively decreasing the bitwidth from high-precision to low-precision during the course of training. 
However, the multi-stage strategy may 
slow down the training.
Inspired by those dropout approaches, we improve the progressive quantization by proposing an efficient single-stage stochastic precision strategy.
Our study shows that this extended scheme is complementary to the proposed joint knowledge distillation approach. 

Recently, Yu \etal proposed slimmable neural networks \cite{yu2019slimmable, yu2019universally} aiming to train a single neural network executable at different widths on the fly at test time, permitting instant and adaptive accuracy-efficiency trade-offs. The core challenge is to sufficiently train all sub-networks.
Specifically, the original slimmable neural networks \cite{yu2019slimmable} proposes to average gradients of different widths without introducing stochastic selection. To improve the optimization of all sub-networks, US-Nets \cite{yu2019universally} randomly samples widths in a certain range and apply averaged gradients back-propagated from the accumulated loss. 
However, our approach aims to solve the notorious difficulty in propagating gradients through a low-precision network due to the non-differentiable quantization function. Inspired by the dropout strategies, we propose to stochastically quantize a portion of the network to low-bit while keeping the other portion full-precision, thus making gradients back-propagate more easily. 
In our method, there is no ``sub-networks" concept. We aim to maximize the performance of the whole quantized network rather than optimizing the mixed-precision network stochastically generated in each iteration sufficiently.
\section{Methods}

In this section, we first introduce the quantization preliminaries in Sec.~\ref{sec:preliminary}. We then describe the progressive quantization schemes in Sec.~\ref{sec:progressive} and explain the stochastic precision approach in Sec.~\ref{sec:stochastic}.
Finally, we elaborate the joint knowledge distillation in Sec.~\ref{sec:mutual}.

\subsection{Problem definition} \label{sec:preliminary}
In this work, we use DoReFa-Net \cite{zhou2016dorefa}\footnote{It should be noted that our proposed method is orthogonal to %
other 
quantization methods.} to quantize both weights and activations. Consider the general case of $k$-bit quantization.  We define the quantization function 
$Q(\cdot)$ as:
\begin{equation}
    z_q = Q(z) = \frac{1}{2^k-1} \cdot {\rm{round}}((2^k-1) \cdot z_r),
\label{eq:quantization1}
\end{equation}
where ${z_r} \in [0,1]$ denotes the normalized full-precision value and ${z_q} \in [0,1]$ denotes the normalized quantized value. With this quantization function, we can define the weight quantization process and the activation quantization process as follows:

\noindent \textbf{Quantization on weights}:
\begin{equation}\label{eq:quan-weigtht}
	{w_q} = 2Q \left( 
	  \frac{{\tanh (w)}}{{2\max (\left| {\tanh (w)} \right|)}} + \frac{1}{2}
	  \right)  - 1.
\end{equation}
In other words, we first use $\frac{{\tanh (w)}}{{2\max (\left| {\tanh (w)} \right|)}} + \frac{1}{2}$ to obtain a normalized version of $w$ and then perform the quantization, where $\tanh(\cdot)$ is adopted to reduce the impact of large values. 

\noindent \textbf{Quantization on activations}:

 Same as \cite{zhou2016dorefa}, we first use a clip function $f(x) = {\rm{clip}}(x,\,0,1)$ to bound the activations to $[0, 1]$. After that, we quantize the activation by applying the quantization function $Q(\cdot)$ on $f(x)$:
\begin{equation} \label{eq:quan-activations}
	{x_q} = Q(f(x)).
\end{equation}

\noindent \textbf{Back-propagation with quantization function}: 

In general, the quantization function is non-differentiable and thus it is impossible to directly apply the back-propagation to train the network. To overcome this issue, we adopt the straight-through estimator \cite{zhou2016dorefa, hubara2016binarized, bengio2013estimating} to approximate the gradients calculation. Formally, we approximate the partial gradient $\frac{{\partial {z_q}}}{{\partial {z_r}}}$ with an identity mapping, namely $\frac{{\partial {z_q}}}{{\partial {z_r}}} \approx 1$.  Accordingly, $\frac{{\partial \ell}}{{\partial {z_r}}}$ can be approximated by 
\begin{equation} \label{eq:STE}
	\frac{{\partial \ell}}{{\partial {z_r}}} = \frac{{\partial \ell}}{{\partial {z_q}}}\frac{{\partial {z_q}}}{{\partial {z_r}}} \approx \frac{{\partial \ell}}{{\partial {z_q}}},
\end{equation}
where $\ell$ is the loss.

\subsection{Progressive quantization} \label{sec:progressive}

\subsubsection{Two-stage optimization}\label{sec:two-stage}
With the straight-through estimator, it is possible to directly optimize the low-precision network. However, the gradient approximation of the quantization function inevitably introduces noisy signals for updating network parameters. Strictly speaking, the approximated gradient may not be the right updating direction. Thus, the training process can be more likely to get trapped at a poor local minimal than training a full-precision model. Applying the quantization function to both weights and activations further worsens the situation.

To %
alleviate 
this training difficulty, we devise a two-stage optimization procedure as follows.
At the first stage, we only quantize the weights of the network while setting the activations to be full-precision. After the converge (or after a certain number of iterations) of this model, we further apply the quantization function on the activations as well and retrain the network. Essentially, the first stage of this method is a related sub-problem of the target one. Compared to the target problem, it is easier to optimize since it only introduces quantization function on weights. Thus, we are more likely to arrive at a good solution for this sub-problem. Then, using it to initialize the target problem may help the network avoid poor local minima which is likely to be encountered if we train the network from scratch.

Let $\Ml^{K}$ be the high-precision model with $K$-bit. We propose to learn a low-precision model $\Ml^{k}$ in a two-stage manner with $\Ml^{K}$ serving as the initial point, where $k<K$. 
 The detailed algorithm is shown in Algorithm \ref{algo:two-stage}.
 
\begin{algorithm}[tb]
	\KwIn{Training data $\{ ({\bx_i},y_i)\}_{i=1}^N$; A $K$-bit precision model $\Ml^K$ with weights $\bWl^K$.}
	\KwOut{A low-precision deep model $\Ml^k$ with weights $\bWl^k$ and activations being quantized into $k$-bit.}
	
	\textbf{Stage 1}: Quantize $\bWl^K$:\\
	\For{ $\mathrm{epoch} = 1,...$ %
	       }
	{               
		\For{ $i = 1,...N$}
		{
			Randomly sample a mini-batch data;\\
			Quantize the weights $\bWl^K$ into $k$-bit by calling some quantization methods with $K$-bit activations\;
		}
	}
	\textbf{Stage 2}: Quantize activations:\\
	Initialize $\bWl^k$ using the converged $k$-bit weights from \textbf{Stage 1} as the starting point; \\
	\For{ $\mathrm{epoch} = 1,...$}
	{
		\For{ $i = 1,...N$}
		{
			Randomly sample a mini-batch data;\\
			Quantize the activations into $k$-bit  by calling some quantization methods while keeping the weights to $k$-bit;
		}
	}
	\caption{Two-stage optimization for $k$-bit quantization}
	\label{algo:two-stage}
\end{algorithm}

\subsubsection{Progressive precision} \label{sec:precision}

The aforementioned two-stage optimization approach suggests the benefits of using a relatively easy-to-optimize problem to find a good initialization. However, separating the quantization of weights and activations is not the only solution to implement the above idea. In this paper, we also propose a second  scheme which progressively lowers the bitwidth of the quantization during the course of network training.  
Specifically, we progressively conduct the quantization from higher precisions to lower precision (e.g., 32-bit $\to$ 16-bit $\to$ 4-bit $\to$ 2-bit). The model of higher precision will be used as the starting point of the relatively lower precision, in analogy with annealing.

Let $\{{b_1},...,{b_n}\}$ be a  sequence precision, where  $b_n<b_{n-1}, ..., b_2<{b_1}$, $b_n$ is the target precision and $b_1$ is set to 32 by default. The whole progressive optimization procedure is summarized in Algorithm~\ref{algo:progressive optimization}.

 Let $\Ml^{k}$ be the low-precision model with $k$-bit and $\Mf$ be the full-precision model. In each step, we propose to learn $\Ml^{k}$, with the solution in the $(i-1)$-th step, denoted by $\Ml^{K}$, serving as the initial point, where $k<K$. 
 
 \begin{algorithm}[bt]
	\KwIn{Training data $\{ ({\bx_j},y_j)\}_{j=1}^N$; A pre-trained 32-bit full-precision  model $\Mf$ as baseline; the precision sequence $\{{b_1},...,{b_n}\}$ where $b_n<b_{n-1}, ..., b_2<{b_1} = 32$.}
	\KwOut{A low-precision deep model $\Ml^{b_n}$.}
	Let $\Ml^{b_1}=\Mf$, where $b_1 = 32$\;	
	\For{ $i = 2,...n$}
	{	
		Let $k = b_i$ and $K=b_{i-1}$\;	
		Obtain $\Ml^{k}$ by calling some quantization methods with $\Ml^{K}$  being the input\;	
	}
	\caption{Progressive precision for accurate CNNs with low-precision weights and activations}
	\label{algo:progressive optimization}
\end{algorithm}

\subsection{Stochastic precision} \label{sec:stochastic}

In the proposed progressive quantization method, we need to gradually quantize the network to low-precision in multi-round training stages. However, the multiple re-initialization and fine-tuning steps may introduce additional computation overhead. To solve this problem, this section develops a single-stage stochastic precision (SP) strategy to improve the training efficiency while enjoying the advantage of the multi-stage progressive quantization. Inspired by the studies that incrementally or stochastically train a certain part of the network \cite{zhou2017incremental, yu2019slimmable}, we propose to incorporate the stochasticity into the progressive training.

The term ``stochastic structure'' means that we %
randomly choose a network structural component, namely, layers, blocks, activations or weights to quantize and keep the rest to be full-precision. The specific scheme is elaborated as follows.

Suppose that we decompose the network $M$ into $Z$ fragments $M=\{{m_1},...,{m_Z}\}$, where ${m_i}$ can be any structure such as a convolutional layer or a residual block.
For each iteration, we intend to partition the fragments into two sets, a low-precision set ${G_q} = \{ {m_{q_1}},...,{m_{q_{N_q}}}\}$ and a full-precision set ${G_r} = \{ {m_{r_1}},...,{m_{r_{N_r}}}\}$, which satisfies the condition:
\begin{equation}
   {G_q} \cup {G_r} = M,\ \mathrm{and}\ {G_q} \cap {G_r} = \emptyset.
\end{equation}
where $N_q$ and $N_r$ are the number of elements in two sets respectively.

In our method, we randomly partition $\Ml$ into $G_q$ and $G_r$. This is implemented by introducing a binary indicator $\bb \in {\mathbb{R}^Z}$ and a stochastic ratio $\delta$.
We randomly set $\bb(i) = 1$ with probability $(1-\delta)$, and if $\bb(i) = 1$ the $i$-th fragment is 
quantized and otherwise is
kept to be full-precision. We linearly decrease $\delta$ to 0 to ensure the whole network being quantized in the end. Note that this procedure implicitly achieves the effect of the incremental quantization \cite{zhou2017incremental} but without the need of multi-round training.

To further increase the randomness in quantizing $m$, we can stochastically choose whether to quantize weights or activations or both of them. This can be implemented by randomly sampling a binary indicator matrix ${\bf{B}} \in {\mathbb{R}^{Z \times 2}}$, where its first column is used to decide whether to quantize the weights in the corresponding fragment and the second column is used to decide whether to quantize activations respectively. 

As a result, $G_q$ can be further partitioned into three subsets $\{{G_{qwa}},{G_{qw}},{G_{qa}}\}$, which represents quantizing both weights and activations, only quantizing weights and only quantizing activations, respectively.
Thus,
SP can share the advantage of the progressive training in Sec.~\ref{sec:two-stage} and Sec.~\ref{sec:precision}. 

Moreover, in Sec.~\ref{exp:different_SP}, we will explore the effect of different structure choices of $m$ as well as the extent of randomness to the final performance.

\begin{algorithm}[tb]
		\KwIn{Training data $\{ {\bx^t},{\by^t}\}$; weights $\bW^t$ ($\bW^0$ = $\bWf$); stochastic ratio $\delta^t$ and decay rate $ \mu$.
		}
		\KwOut{Updated parameters $\bW^{t+1}$; stochastic ratio $\delta ^ {t+1}$.}
		Partition $M$ into $Z$ fragments $\{{m_1},...,{m_Z}\}$; \\
		\If{ $\delta^t > 0$}
		{ Obtain the binary indicator matrix $\bB^t$ via uniform sampling with probability $\delta^t$; \\
		Partition the network $M$ into the quantized set $G_q = \{ {G_{qwa}^t},{G_{qw}^t},{G_{qa}^t}\}$ and the full-precision set $G_r^t$ according to ${\bf{B}}^t$; \\
		Obtain the corresponding mixed-precision parameter set $\widetilde \bW^t = \{ Q({\bW^t_{qwa}}),Q({\bW^t_{qw}}),{\bW^t_{qa}},{\bW^t_r}\}$ ;
	    }

	    \Else{$\widetilde \bW^t = Q(\bW^t) $;}
	  
		${{\mathbf{\widetilde{y}}}^t} = \mathrm{Forward}({\bx^t},\widetilde \bW^t)$; \\
		Compute the loss $\cL({\by^t},\widetilde \by^t) $; \\
		$\frac{{\partial \cL}}{{\partial {\widetilde \bW^t}}} = \mathrm{Backward}(\frac{{\partial \cL}}{{\partial {{\mathbf{\widetilde{y}}}^t}}},{\widetilde \bW^t})$; \\
	    Update parameters $\bW^{t+1}$ using STE defined in Eq. (\ref{eq:STE}) and some proper optimizers; \\
	    $\delta^{t+1} = \delta^t  - \mu$;
		\caption{Stochastic precision training algorithm.}
		\label{algo:stochastic}
	\end{algorithm}

\subsection{Joint knowledge distillation on quantization}\label{sec:mutual}

\begin{figure*}[tb]
	\centering
	\resizebox{0.65\linewidth}{!}
	{
		\begin{tabular}{c}
			\includegraphics{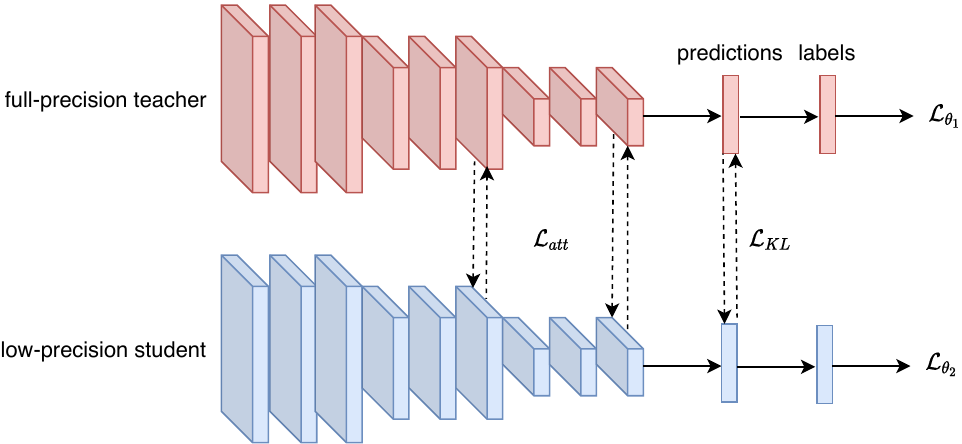}
		\end{tabular}
	}
	\caption{Demonstration of the guided training strategy.
	Dashed lines show the guidance  loss.
	}
	\label{fig:knowledge_transfer}
\end{figure*}

The third approach proposed
here 
is inspired by the success of using information distillation ~\cite{romero2014fitnets, hinton2015distilling, parisotto2016actor, zagoruyko2016paying, ba2014deep} to train a relatively shallow network. Specifically, these methods usually use a teacher model (usually a pretrained deeper network) to provide guided signal for the shallower network. Following this motivation, we propose to train the low-precision network alongside another guidance network. Unlike the work in \cite{romero2014fitnets, hinton2015distilling, parisotto2016actor, zagoruyko2016paying, ba2014deep}, the guidance network shares the similar architecture as the target network but is pretrained with full-precision weights and activations. 

However, a pre-trained model may not be necessarily optimal or may not be suitable for quantization. As a result, directly using a fixed pretrained model to guide the target network may not produce the best guidance signals. To mitigate this problem, we do not fix the parameters of a pretrained full-precision network.

By using the guidance training strategy, we assume that there exist some full-precision models with good generalization performance, and an accurate low-precision model can be obtained by directly performing the quantization on those full-precision models. In this sense, the feature maps of the learned low-precision model should be close to that obtained by directly performing quantization on the full-precision model. To achieve this, essentially, in our learning scheme, we can jointly train the full-precision and low-precision models. This allows these two models adapt to each other. We even find by doing so the performance of the full-precision model can be slightly improved in some cases.  

\begin{algorithm}[b]
	\KwIn{Training data $\{ (\bx_i,y_i)\}_{i=1}^N$; A pre-trained 32-bit full-precision model $\Mf$; A $k$-bit precision model $\Ml^k$.}
	\KwOut{A low-precision deep model $\Ml^k$ with weights and activations being quantized into $k$ bits.}
	Initialize $\Ml^k$ based on $\Mf$;\\
	\For{ $\mathrm{epoch} = 1,...$ %
	        }
	{		
		\For{ $i = 1,...N$}
		{
			Randomly sample a mini-batch data;\\
			Quantize the weights $\bWl$  and activations into $k$-bit by minimizing $\cL_2(\bWl)$\;
			Update $\Mf$ by minimizing $\cL_1(\bWf)$\;
		}
	}
	\caption{Guided training with a full-precision network for $k$-bit quantization}
	\label{algo:one-mutual learning}
\end{algorithm}

Formally, let $\bWf$ and $\bWl$ be the weights of the full-precision model and low-precision model, respectively. 
Let $\mu (\bx; \bWf)$ and $\nu (\bx;\bWl)$ be the nested feature maps (i.e., activations) of the full-precision model and low-precision model, respectively. To create the guidance signal, we may require that the nested feature maps from the two models should be similar. However,  $\mu (\bx;\bWf)$ and $\nu (\bx;\bWl)$  is usually not directly comparable since one is full-precision and the other is low-precision.
To link these two models,  we can directly quantize the weights and activations of the full-precision model. For simplicity, we denote the quantized feature maps by  $Q(\mu (\bx;\bWf))$. Thus, $Q(\mu (\bx;\bWf))$ and  $\nu (\bx;\bWl)$ will become comparable. 
Inspired by the attention transfer method \cite{zagoruyko2016paying}, we propose to apply attention matching at a set of $\mathcal{T}$ transfer points within a network, the constraint can be expressed as:
\begin{equation}  \label{eq:atten}
   \cL_{att}(\bWf,\bWl) =  \sum\limits_{i = 1}^N\sum\limits_{j = 1}^\mathcal{T} 
   \left\lVert
   \frac{\mathcal{A}^j_S}{\parallel\mathcal{A}^j_S\parallel_2} - \frac{\mathcal{A}^j_T}{\parallel\mathcal{A}^j_T\parallel_2}
   \right\rVert,
\end{equation}
where $\mathcal{A}^j_S$ and $\mathcal{A}^j_T$ are the sum of the absolute values across the channel dimension of feature maps $\nu (\bx_i;\bWl)$ and $Q(\mu (\bx_i;\bWf))$, respectively.

Similar to \cite{hinton2015distilling}, we can also employ the posterior probability as the guidance signal. Let $\bpf$ and $\bpl$ be the full-precision teacher network and low-precision student network predictions, respectively. To measure the correlation between
the two distributions, we employ the Kullback–Leibler (KL) divergence:
\begin{equation} \label{eq:pos}
{\cL_{\rm{KL}}}(\bpf|\bpl) = \sum\limits_{i = 1}^N {\bpf(\bx_i)} \log \frac{{{\bpf}({\bx_i})}}{{{\bpl}(\bx_i)}}.
\end{equation}

Finally, let ${\cL_{{\theta _1}}}$ and ${\cL_{{\theta _2}}}$ be the cross-entropy classification losses for the full-precision and low-precision model, respectively. The guidance losses in Eqs. (\ref{eq:atten}) and (\ref{eq:pos}) will be added to ${\cL_{{\theta _1}}}$ and ${\cL_{{\theta _2}}}$, respectively, resulting in the final objectives for the two networks, namely
\begin{equation}
{\cL_1}(\bWf) = \alpha_1 {\cL_{{\theta _1}}} + \beta  {\cL_{\rm{KL}}}(\bpl|\bpf) + \gamma \cL_{att}(\bWf,\bWl),
\end{equation}
and 
\begin{equation}
{\cL_2}(\bWl) = \alpha_2 {\cL_{{\theta _2}}} + \beta  {\cL_{\rm{KL}}}(\bpf|\bpl) +  \gamma \cL_{att}(\bWf,\bWl),
\end{equation}
where $\{\alpha_1, \alpha_2\}$, $\beta$ and $\gamma$ are the balancing hyper-parameters. We share $\beta$ and $\gamma$ for these two objectives.

In the learning procedure, both $\bWf$ and $\bWl$ will be updated by minimizing $\cL_1(\bWf)$ and $\cL_2(\bWl)$ separately, using a mini-batch stochastic gradient descent method. The detailed algorithm is shown in Algorithm \ref{algo:one-mutual learning}. A high-bit precision model $\Ml^K$ is used as an initialization of $\Ml^k$, where $K>k$. Specifically, for the full-precision model, we have $K=32$. Relying on $\Mf$, the weights and activations of $\Ml^k$ can be initialized respectively.

Note that the training process of the two networks are different.
When updating $\bWl$ by minimizing $\cL_2(\bWl)$, we use the full-precision model as initialization and apply STE to fine-tune the model. When updating $\bWf$ by minimizing $\cL_1(\bWf)$, we use conventional forward-backward propagation to fine-tune the model.

\subsection{Remarks on the proposed methods}
The proposed three approaches tackle the difficulty in training a low-precision model with different strategies. They can be applied independently. However, it is also possible to combine them together. For example, we can apply the progressive precision to any step in the two-stage approach; we can also apply the joint knowledge distillation to any step in the progressive quantization; we can combine stochastic precision with the joint knowledge distillation approach.
Detailed analysis on possible combinations will be empirically evaluated in the experiment section.

\section{Experiments}  \label{exp:experiment}

\noindent \textbf{Datasets and models.} To investigate the performance of the proposed methods, we conduct experiments on CIFAR-100~\cite{krizhevsky2009learning} and ImageNet~\cite{russakovsky2015imagenet}. We employ ResNet~\cite{he2016deep}, PreResNet~\cite{he2016identity} and AlexNet~\cite{krizhevsky2012imagenet} for experiments.
We use a variant of the AlexNet structure by removing dropout layers and add batch normalization after each convolutional layer and fully-connected layer. This structure is widely used in previous works~\cite{zhou2016dorefa, zhu2016trained}.

\noindent \textbf{Comparison methods.} To justify the effectiveness of the proposed approaches, we conduct experiments on various representative quantization approaches, including uniform fixed-point approach DoReFa-Net~\cite{zhou2016dorefa}, non-uniform fixed-point method LQ-Net~\cite{zhang2018lq}, as well as binary neural network approaches BiReal-Net~\cite{Liu_2018_ECCV} and Group-Net~\cite{zhuang2019structured}.
The ``Baseline'' in all experiments means that we quantize the model using DoReFa-Net \cite{zhou2016dorefa}, which is defined in Sec.~\ref{sec:preliminary}.
We define ``TS'', ``PP'', ``SP'' and ``KD'' to represent two-stage optimization in Sec.~\ref{sec:two-stage}, progressive precision in Sec.~\ref{sec:precision}, stochastic precision in Sec.~\ref{sec:stochastic} and joint knowledge distillation in Sec.~\ref{sec:mutual}, respectively.

\noindent \textbf{Implementation details.} As in \cite{rastegari2016xnor, Cai_2017_CVPR, zhou2016dorefa, zhuang2018towards, zhuang2019structured}, we quantize the weights and activations of all convolutional layers except that the first and the last layers are kept in full-precision.
However, we also quantize all the layers so that the model contains complete fixed-point operations and we label this case with a * symbol.
In all ImageNet experiments, training images are resized to $256 \times 256$, and a $224 \times 224$ crop is randomly sampled from an image or its horizontal flip, with the per-pixel mean subtracted. We do not use any further data augmentation in our implementation. We use a simple single-crop testing for standard evaluation. No bias term is utilized.

Without loss of generality, we finetune from the pretrained full-precision model and set the initial learning rate for the quantized model to 0.005. We train a maximum 30 epochs, and decay the learning rate by 10 at the 15-th and 25-th epoch. We use SGD for optimization, with a batch size of 256, a momentum of 0.9 and a weight decay of  1e-4. More specific hyperparameters are provided in each subsection.
Our implementation is based on PyTorch.

\subsection{Effect of progressive quantization} \label{exp:relaxed}
In this part, we explore the effect of the proposed progressive quantization methods.     

\subsubsection{Effect of the two-stage optimization} \label{exp:two-step}
\begin{figure}[b]
	\centering
	\resizebox{0.8\linewidth}{!}
	{
		\begin{tabular}{c}
			\includegraphics{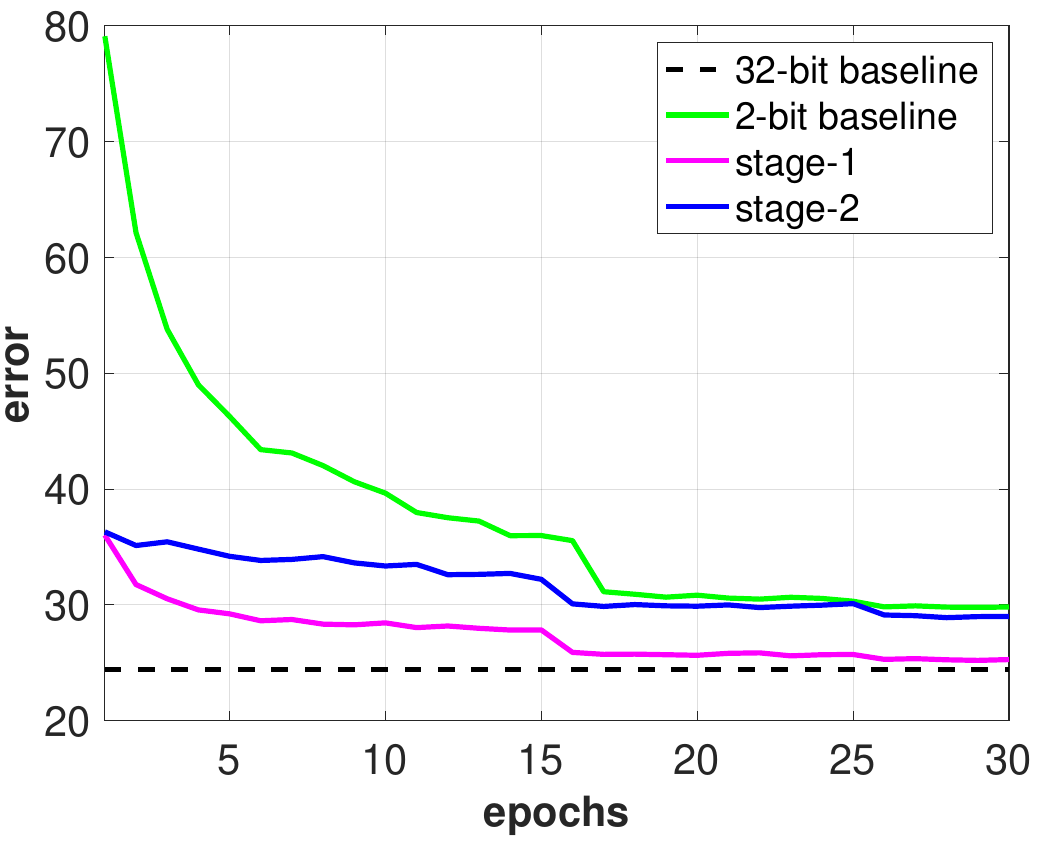}
		\end{tabular}
	}
	\caption{The two-stage training approach on ResNet-50.}
	\label{fig:two_step}
\end{figure}

We analyze the effect of each stage in the two-stage approach in Figure~\ref{fig:two_step}.
We take the 2-bit ResNet-50 on ImageNet as an example. In Figure~\ref{fig:two_step}, step-1 has the minimal loss of accuracy. As for the step-2, although it incurs an apparent accuracy decrease in comparison with that of the step-1, its accuracy is consistently better than the results of the baseline in every epoch. This illustrates that progressively seeking for the local minimum point is crucial for final better convergence, which proves the effectiveness of this simple mechanism. 

\begin{table}[!htb]
	\centering
	\caption{Accuracy (\%) of different comparing methods on the ImageNet validation set.}
	\scalebox{0.9}
	{
		\begin{tabu}{c c | c | c c}
		  Precision & model &	method &top-1 acc. &top-5 acc. \\
			\tabucline[1pt]{-}
		\multirow{4}{*}{2W, 2A} &\multirow{4}{*}{ResNet-50} &Baseline  &70.19  &89.15  \\ 
		  &	& Baseline + TS  &70.92  &90.03  \\
		  &	& Baseline + PP &70.78  &89.98 \\
		  & & Baseline + TS + PP &\bf{71.13}  &\bf{90.12}  \\
		  \hline
	     \multirow{4}{*}{4W, 4A} &\multirow{4}{*}{ResNet-50*} &Baseline  &75.11  &75.70  \\ 
		  &	& Baseline + TS  &75.32  &91.93  \\
		  &	& Baseline + PP &75.38  &91.77 \\
		  & & Baseline + TS + PP &\bf{75.50}  &\bf{92.01} \\
		  \hline
		\multirow{4}{*}{2W, 2A} &\multirow{4}{*}{ResNet-50*} &Baseline  &67.68  &70.00  \\ 
		  &	& Baseline + TS  &69.22  &87.03  \\
		  &	& Baseline + PP &68.79  &86.90 \\
		  & & Baseline + TS + PP &\bf{69.43}  &\bf{87.01} \\
		  \hline
		 \multirow{4}{*}{4W, 4A} &  \multirow{4}{*}{AlexNet*} &Baseline  &56.17  &79.35 \\
		  &	&Baseline + TS &57.66  &81.03 \\ 
		  &	&Baseline + PP &57.47  &80.80 \\
		  & &Baseline + TS + PP &\bf{57.83}  &\bf{80.85} \\ 
		  \hline
		 \multirow{4}{*}{2W, 2A} &  \multirow{4}{*}{AlexNet*} &Baseline  &48.26  &71.56  \\
		  &	&Baseline + TS &50.67  &74.92 \\ 
		  &	&Baseline + PP &50.29  &74.80 \\
		  & &Baseline + TS + PP &\bf{50.94}  &\bf{74.93} \\
	\end{tabu}}
	\vspace{-1.0em}
	\label{tab:exp_twostep_progressive}
\end{table}

\subsubsection{Effect of the progressive precision strategy}  \label{exp:progressive}
\begin{figure}[!htb]
	\centering
	\resizebox{0.8\linewidth}{!}
	{
		\begin{tabular}{c}
			\includegraphics{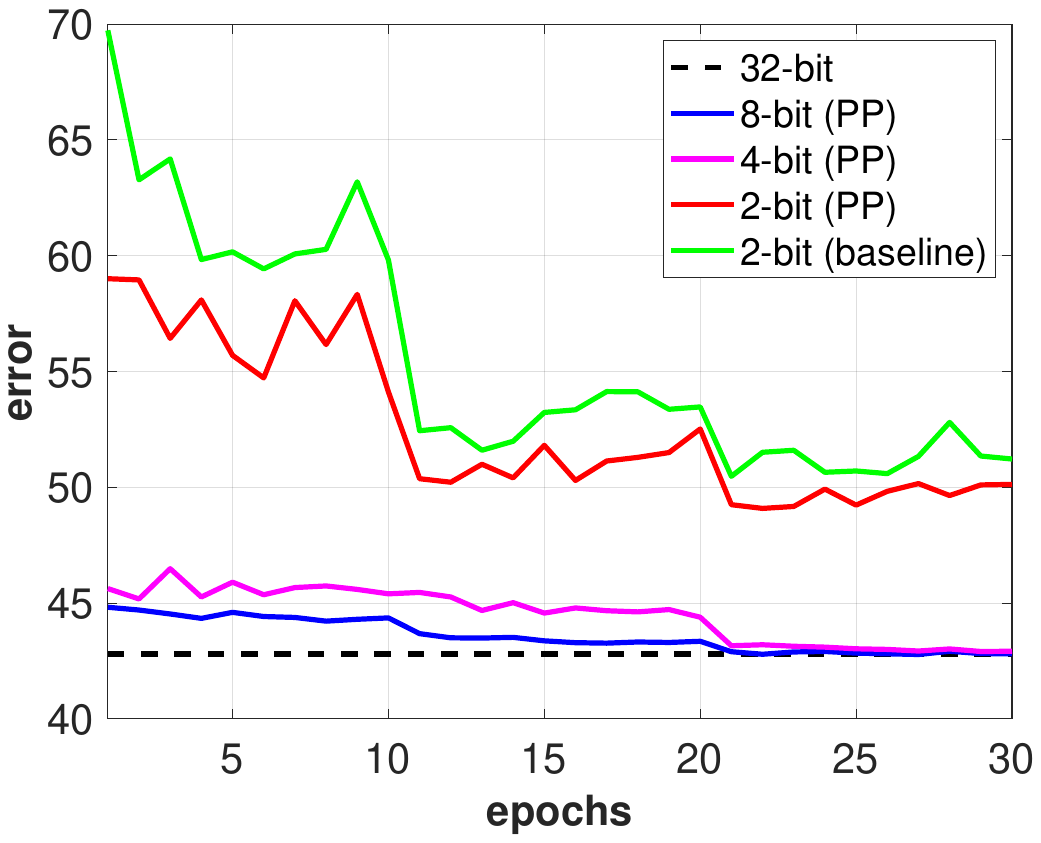}
		\end{tabular}
	}
	\caption{The progressive training approach on AlexNet*.}
	\label{fig:progressive}
	\vspace{-1.0em}
\end{figure}

What is more, we also separately explore the progressive precision effect on the final performance. 
In this experiment, we apply AlexNet and ResNet-50 on the ImageNet dataset. 
We continuously quantize both weights and activations simultaneously from 32-bit$\to$8-bit$\to$4-bit$\to$2-bit and explicitly illustrate the accuracy change process for each precision in Figure~\ref{fig:progressive}. The quantitative results are also reported in Table~\ref{tab:exp_twostep_progressive}. From the figure, we find that for 8-bit and 4-bit, the low-bit model has no accuracy loss with respect to the full-precision model. However, when quantizing from 4-bit to 2-bit, we can observe a significant accuracy drop. 
Despite this, we still observe $2.0\%$ relative improvement by comparing the Top-1 accuracy over the 2-bit baseline, which proves the effectiveness of the proposed strategy. It is worth noting that the accuracy curves become more unstable when quantizing to the lower bit. This phenomenon is reasonable because the quantized value will change more frequently during the training process when the bitwidth is reduced. 

\subsection{Effect of the stochastic precision}   \label{exp:stochastic}
In this subsection, we further explore the effect of the stochastic precision strategy on general quantization approaches. The stochastic ratio $\delta$ is initialized to 0.5 and linearly decayed to 0 at the 20-th epoch. We train a maximum 40 epochs and decay the learning rate by 10 at the 25-th and 35-th epochs. Other hyperparameters are set to default.
The default structure of the fragment $m$ is a residual block and we stochastically quantize weights and activations in all cases unless special explanations.
The results are reported in Table~\ref{tab:exp_sp}. By combining the baseline methods with \emph{SP}, we find an apparent performance increase compared with the baselines in all cases. During training, we stochastically keep a portion of the network to full-precision and update by the standard gradient-based method. This strategy shares the similar spirit with the progressive quantization to relax the discrete quantizer effectively. Moreover, the proposed stochastic strategy only requires one training stage without fine-tuning the model in many training rounds.

\begin{table}[!htb]
	\centering
	\caption{Accuracy (\%) of different comparing methods with SP on the ImageNet validation set. Experiments are repeated for 3 times and we report the results with mean and standard deviation.}
	\scalebox{1.0}
	{
		\begin{tabu}{c | c | c c}
		    model &	method &top-1 acc. &top-5 acc. \\
			\tabucline[1pt]{-}
			\multirow{2}{*}{ResNet-50} & DoReFa-Net (2-bit)  &70.19  &89.15  \\ 
			& DoReFa-Net + SP  &\bf{72.23$\pm$0.05}  & \bf{90.78$\pm$0.10}  \\
			\hline
		    \multirow{2}{*}{ResNet-50} & LQ-Net (3-bit) &74.23  &91.63 \\
			& LQ-Net + SP &\bf{75.14$\pm$0.04}  &\bf{92.33$\pm$0.09} \\ 
			\hline 
			\multirow{2}{*}{ResNet-18} & BiReal-Net &56.43  &79.52 \\
			& BiReal-Net + SP & \bf{58.81$\pm$0.05}  &\bf{81.24$\pm$0.12} \\
			\hline
			\multirow{2}{*}{ResNet-18} & GroupNet (5 bases) &64.82  &85.72 \\
			& GroupNet + SP &\bf{65.89$\pm$0.06}  &\bf{86.30$\pm$0.10} \\
			
	\end{tabu}}
	\label{tab:exp_sp}
\end{table}

\begin{figure}[!htb]
	\centering
	\resizebox{0.8\linewidth}{!}
	{
		\begin{tabular}{c}
			\includegraphics{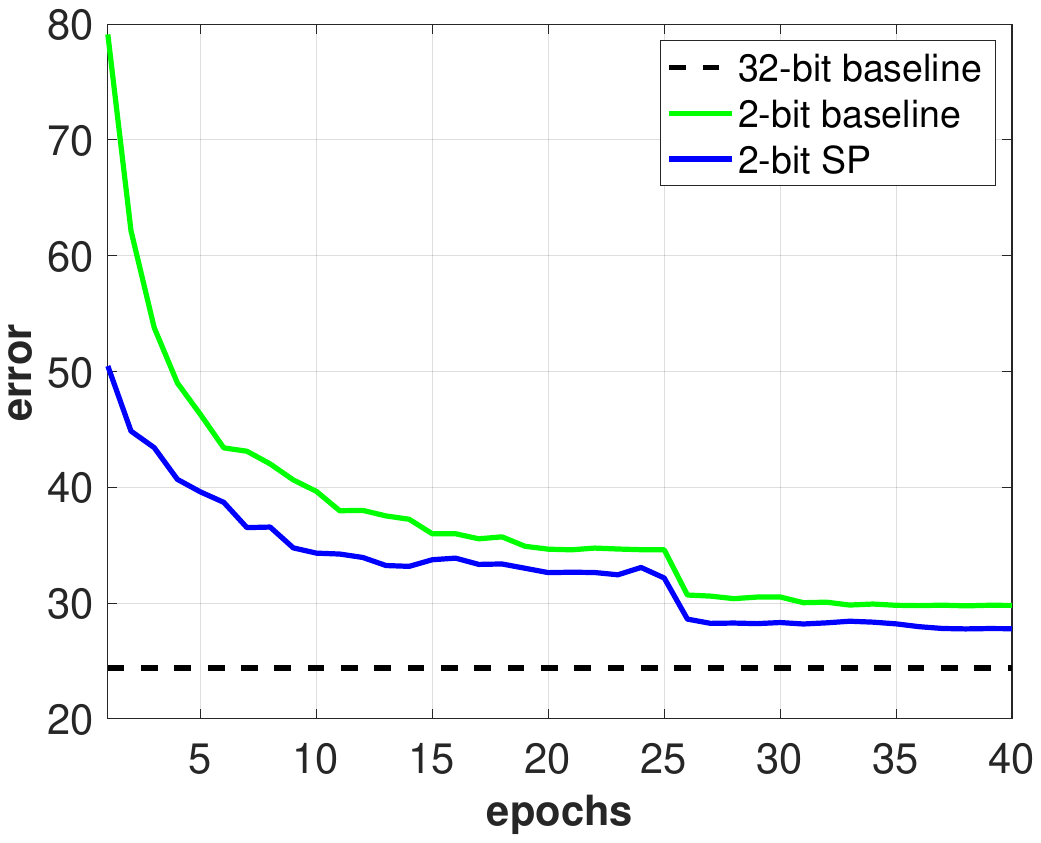}
		\end{tabular}
	}
	\caption{The stochastic precision training approach on
	ResNet-50.}
	\label{fig:SP}
\end{figure}

\subsubsection{Effect of different SP policies}   \label{exp:different_SP}
We further explore the influence of different choices of the fragment $m$ described in Sec.~\ref{sec:stochastic} as well as the extent of randomness.
We treat GroupNet as our baseline approach and utilize 5 binary bases. The results are reported in Table~\ref{tab:exp_stochastic}.
We explore two different structures of $m$, including one convolutional layer and one residual block which corresponds to \emph{layerdrop} and \emph{blockdrop} respectively. We further incorporate the randomness of quantizing weights and activations into $m$ and is denoted by \emph{W/A}. From the results, we observe that all the four cases show improved performance compared with the baseline, which justifies adding randomness is a general way for relaxing the low-precision network training. 
By comparing the result of \emph{layerdrop+W/A} with \emph{layerdrop}, we observe performance drop with the increase of randomness. However, \emph{blockdrop+W/A} performs slightly better than \emph{blockdrop}. This shows that adding excessive stochasticity can make the gradient updating direction deviate while appropriate extent of randomness can relax the non-differentiable problem to facilitate optimization. Moreover, the accuracy of \emph{layerdrop} and \emph{blockdrop} are very close, which shows that the structure of $m$ is not sensitive to the final performance.

\begin{table}[!htb] 
	\centering
	\caption{Accuracy (\%) of different stochastic policies on the ImageNet validation set.}
	\scalebox{0.9}
	{
		\begin{tabu}{c | c | c c}
		    model &	method &top-1 acc. &top-5 acc. \\
			\tabucline[1pt]{-}
			\multirow{5}{*}{ResNet-18} & GroupNet (5 bases) &64.8  &85.7  \\ 
		    & GroupNet + blockdrop &65.6  &86.3 \\
            & GroupNet + layerdrop &65.7  &86.5 \\
            & GroupNet + blockdrop + W/A &\bf{65.9}  &\bf{86.6}  \\
            & GroupNet + layerdrop + W/A &65.0  &86.1  \\
	\end{tabu}}
	\label{tab:exp_stochastic}
\end{table}

\subsection{Effect of the joint knowledge distillation on quantization} \label{exp:mutual}

To investigate the effect of the joint knowledge distillation approach explained in Sec.~\ref{sec:mutual}, we explore four different training schemes to obtain a low-precision student network.

\subsubsection{Joint fine-tuning of the low-precision student and the full-precision teacher}  \label{exp:joint}

In this scheme, both networks are primed with corresponding full-precision pretrained weights as initialization and are jointly optimized. We explore two network structures, including PreResNet and ResNet. When using a certain student network $\Ml$, we use the teacher network $\Mf$ to have either the same or larger depth. The results are reported in Table~\ref{tab:joint_ResNet} and Table~\ref{tab:joint_PreResNet}. 
The initial learning rates for $\Ml$ and $\Mf$ are set to be 0.005 and 0.001, respectively. The balancing 
hyperparameters $\{\alpha_1, \alpha_2\}=\{1, 0.5\}$, $\beta=0.5$ and $\gamma=50$. Other hyperparameters are set to default.

\noindent\textbf{Discussion.} From the results, we observe that all our low-precision models surpass the corresponding baselines. It justifies that $\Mf$ can provide useful auxiliary supervision to assist the convergence of $\Ml$. Moreover, the relative improvement with ResNet is larger than that with PreResNet. To highlight, the relative Top-1 improvement w.r.t.\  2-bit ResNet-50 is 1.77\% while the PreResNet-50 counterpart is 0.71\%. 
This phenomenon can be attributed that quantized ResNet is more difficult to be optimized since the skip connections are also quantized which blocks layers later in the network to access information gained in earlier layers. In this scenario, $\Mf$ can effectively ease the training of $\Ml$ by adapting knowledge to each other. We can also justify that keeping the skip connections to high-precision is important to maintain the performance of the low-precision network similar to~\cite{Liu_2018_ECCV, park2018precision}.

Moreover, we can come to an assumption that the distillation process becomes more effective when the low-precision network is more difficult to train.
This assumption can be further proved by the experiments in Sec.~\ref{exp:scratch}. 

In Table~\ref{tab:joint_PreResNet}, we experiment with PreResNet-18 which is paired with various teacher networks but with deeper layers. However, the benefit of using a deeper network saturates at some points. For example, the final trained accuracy of 2-bit PreResNet-18 model paired with PreResNet-50 is only 0.04\% higher than that obtained by pairing the PreResNet-34 network. 

With the simple DoReFa-Net uniform quantization strategy, we can achieve  comparable or even higher accuracy compared with the full-precision model using 4-bit precision. It means that we can deploy the 4-bit model in hardware devices with no loss of accuracy which would greatly save memory bandwidth and power consumption.  

\begin{figure}[!htb]
	\centering
	\resizebox{0.8\linewidth}{!}
	{
		\begin{tabular}{c}
			\includegraphics{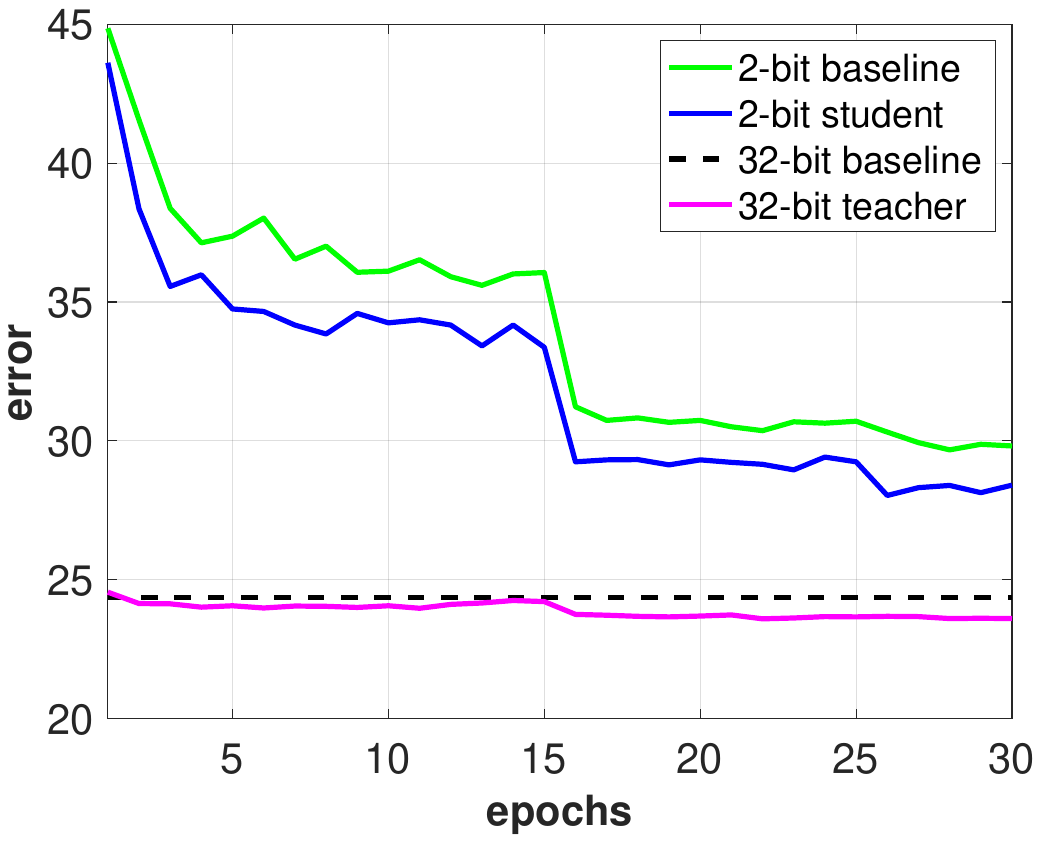}
		\end{tabular}
	}
	\caption{Both student and teacher are fine-tuned from the pretrained models. We use ResNet-50 as illustration.}
	\label{fig:kl_finetune}
\end{figure}

Interestingly, we also observe that the full-precision teacher can also be improved by learning together with the student. We plot the convergence curves in Figure~\ref{fig:kl_finetune}. We can observe that the teacher’s performance drops at the beginning epochs due to inaccurate gradients from the student. During optimization, the
student network serves as a regularizer for the teacher network which can even surpass the pretrained baseline.

\begin{table*}[tb]
	\centering
	\caption{Accuracies of the quantized ResNet using joint training approach and finetuning. W and A refer to the bitwidth of weights and activations, respectively. Experiments are repeated for 3 times and we report the results with mean and standard deviation.}
	\scalebox{0.9}
	{
	\begin{tabular}{cccccccc}
		\toprule
		Precision & & \tabincell{c}{ResNet-18 \\ Baseline} & \tabincell{c} {ResNet-18 \\ with ResNet-18}  & \tabincell{c} {ResNet-34 \\ Baseline} &
		\tabincell{c} {ResNet-34 \\ with ResNet-34}  &\tabincell{c} {ResNet-50 \\ Baseline} & \tabincell{c} {ResNet-50 \\ with ResNet-50}  \\ \midrule
	    \multirow{2}{*}{32W, 32A} &Top-1\% &69.75  &- &73.21 &- &75.64 &-  \\ 
	     &Top-5\% &89.01  &- &91.40 &- &92.25 &- \\ \midrule
	    \multirow{2}{*}{4W, 4A} &Top-1\% &69.47$\pm$0.04 &\bf{70.18$\pm$0.04}  &71.31$\pm$0.03 &\bf{73.08$\pm$0.05} &74.50$\pm$0.03 &\bf{75.67$\pm$0.06}  \\ 
	     &Top-5\% &88.80$\pm$0.09  &\bf{90.20$\pm$0.12}  &90.08$\pm$0.11 &\bf{91.53$\pm$0.10} &91.46$\pm$0.14 &\bf{92.19$\pm$0.13} \\ \midrule
	     
	    \multirow{2}{*}{2W, 2A} &Top-1\% &64.67$\pm$0.04 &\bf{65.58$\pm$0.05}  &68.17$\pm$0.05  &\bf{69.20$\pm$0.06} &70.19$\pm$0.04 &\bf{71.96$\pm$0.06} \\ 
	     &Top-5\% &85.78$\pm$0.10 &\bf{86.44$\pm$0.14} &88.05$\pm$0.09 &\bf{89.09$\pm$0.13} &89.15$\pm$0.08 &\bf{90.63$\pm$0.16} \\
		\bottomrule
	\end{tabular}}
	\label{tab:joint_ResNet}
\end{table*}

\begin{table*}[ht]
	\centering
	\caption{Accuracies of the quantized PreResNet using joint training approach and finetuning.  Experiments are repeated for 3 times and we report the results of mean and standard deviation.}
	\scalebox{0.758}
	{
	\begin{tabular}{cccccccccc}
		\toprule
		Precision & &\tabincell{c}{PreResNet-18 \\ Baseline} & \tabincell{c} {PreResNet-18 \\ with PreResNet-18} & \tabincell{c} {PreResNet-18 \\ with PreResNet-34}  & \tabincell{c} {PreResNet-18 \\ with PreResNet-50} &\tabincell{c} {PreResNet-34 \\ Baseline} & 
		\tabincell{c} {PreResNet-34 \\ with PreResNet-34}  & \tabincell{c} {PreResNet-50 \\ Baseline}
		& \tabincell{c} {PreResNet-50 \\ with PreResNet-50}  \\ \midrule
	    \multirow{2}{*}{32W, 32A} &Top-1\% &69.95  &- &-  &- &73.53 &- &76.11 &-  \\ 
	     &Top-5\% &89.21 &-  &-  &- &91.30 &- &92.81  &- \\ \midrule
	    \multirow{2}{*}{4W, 4A} &Top-1\% &69.81$\pm$0.04  &\bf{70.12$\pm$0.05}  &- &- &73.57$\pm$0.03 &\bf{73.90$\pm$0.03}  &75.92$\pm$0.02  &\bf{76.62$\pm$0.04}  \\ 
	     &Top-5\% &89.04$\pm$0.09 &\bf{89.57$\pm$0.09} &- &- &91.35$\pm$0.08 &\bf{91.62$\pm$0.10} &92.82$\pm$0.07  &\bf{93.14$\pm$0.09} \\ \midrule
	    \multirow{2}{*}{2W, 2A} &Top-1\% &64.51$\pm$0.05  &65.67$\pm$0.06  &65.82$\pm$0.05 &\bf{65.86$\pm$0.04} &69.31$\pm$0.05 &\bf{70.26$\pm$0.04}  &71.20$\pm$0.04  &\bf{71.91$\pm$0.05}  \\ 
	     &Top-5\% &85.85$\pm$0.10 &86.80$\pm$0.11 &\bf{86.84$\pm$0.10}  &86.73$\pm$0.12  &88.92$\pm$0.11  &\bf{89.64$\pm$0.10}  &90.18$\pm$0.10  &\bf{90.53$\pm$0.12} \\	
		\bottomrule
	\end{tabular}}
	\label{tab:joint_PreResNet}
\end{table*}

\subsubsection{Learning from scratch vs.\ fine-tuning} \label{exp:scratch}

In this scheme, we train a low-precision student from scratch given a pretrained full-precision teacher network. During training, both of the models are mutually updated. The initial learning rates for student and teacher are set to be 0.1 and 0.001, respectively. We train a maximum 80 epochs with SGD, and the learning rate is decayed by 10$\times$ at epochs 30, 50, 60 and 70. We use the batch size of 256.

\begin{figure}[!htb]
	\centering
	\resizebox{0.8\linewidth}{!}
	{
		\begin{tabular}{c}
			\includegraphics{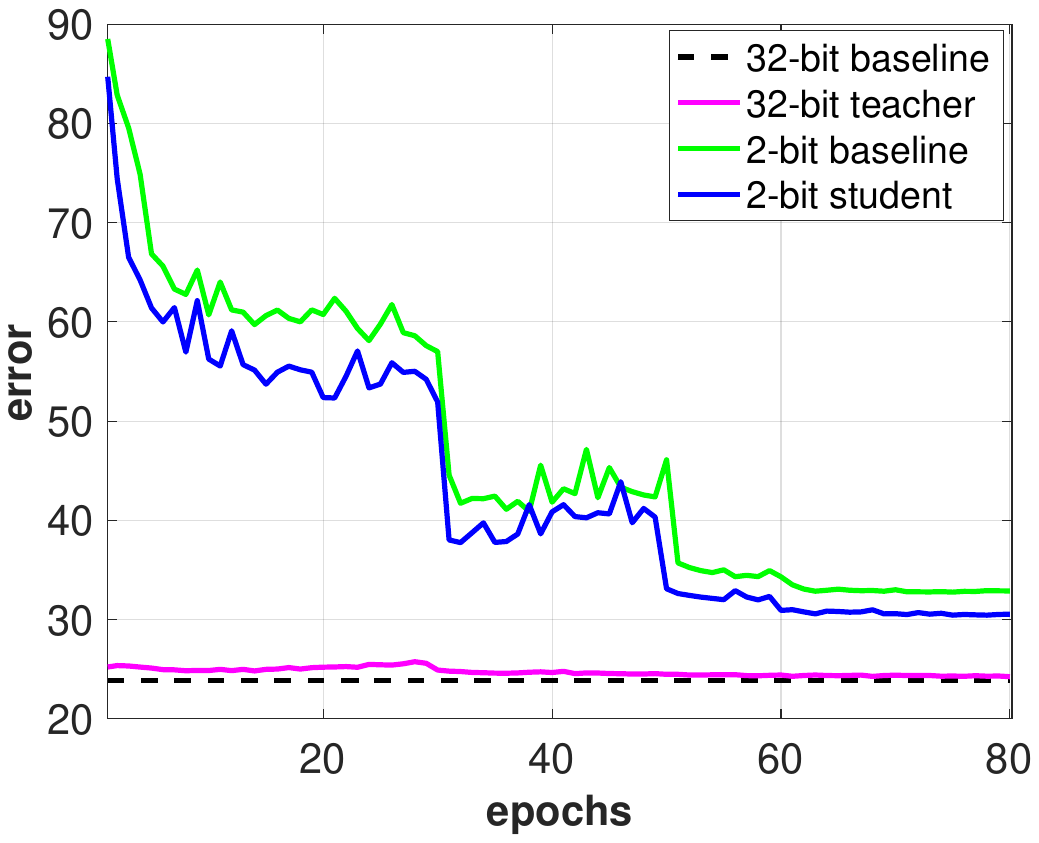}
		\end{tabular}
	}
	\caption{Student is learnt from scratch while teacher is fine-tuned. PreResNet-50 is used here.}
	\label{fig:scratch_kl}
\end{figure}

The results are reported in Table~\ref{tab:scratch}. From the results, we can summarize two instructive statements.
\begin{itemize}
    \item 
The relative improvement of KD is more apparent than those that are from fine-tuning. For instance, with 2-bit representations, the relative improvement for PreResNet-50 is 2.40\% while the fine-tuning counterpart is 0.71\% in Table~\ref{tab:joint_PreResNet}.
This is reasonable since learning from scratch is more challenging than fine-tuning and the auxiliary guidance from the teacher has more affects. 
\item
Fine-tuning performs steadily better than learning from scratch. It shows that the pretrained full-precision model can serve as an important 
initialization.

\end{itemize}
We also %
show
the convergence curves in Figure~\ref{fig:scratch_kl}. 

\begin{table*}[!tb]
	\centering
	\caption{The accuracy of the quantized PreResNet using the joint training approach, which is learnt from scratch. Experiments are repeated for 3 times and we report the results with mean and standard deviation.}
	\scalebox{1.0}
	{
	\begin{tabular}{cccccccc}
		\toprule
		Precision & & \tabincell{c}{PreResNet-18 \\ Baseline} & \tabincell{c} {PreResNet-18 \\ with PreResNet-18} & \tabincell{c} {PreResNet-34 \\ Baseline}  & \tabincell{c} {PreResNet-34 \\ with PreResNet-34} &
		\tabincell{c} {PreResNet-50 \\ Baseline}  & \tabincell{c} {PreResNet-50 \\ with PreResNet-50} \\ \midrule
	    \multirow{2}{*}{32W, 32A} &Top-1\% &69.95  &-  &73.53  &- &76.11 &-  \\ 
	     &Top-5\% &89.21 &-  &91.30 &-  &92.81 &- \\ \midrule
	    \multirow{2}{*}{4W, 4A} &Top-1\%  &67.85$\pm$0.08  &\bf{69.29$\pm$0.10}  &71.46$\pm$0.13 &\bf{73.05$\pm$0.14} &73.82$\pm$0.12  &\bf{75.42$\pm$0.11}  \\ 
	     &Top-5\% &88.15$\pm$0.16 &\bf{88.84$\pm$0.15}  &90.06$\pm$0.14 &\bf{91.01$\pm$0.16} &91.53$\pm$0.13 &\bf{92.82$\pm$0.15} \\ \midrule
	    \multirow{2}{*}{2W, 2A} &Top-1\% &62.54$\pm$0.12 &\bf{65.08$\pm$0.11}  &66.57$\pm$0.13 &\bf{68.69$\pm$0.14} &67.15$\pm$0.13 &\bf{69.55$\pm$0.12}  \\ 
	     &Top-5\% &84.47$\pm$0.16 &\bf{86.21$\pm$0.18}  &87.18$\pm$0.16  &\bf{88.52}$\pm$0.20 &87.74$\pm$0.16 &\bf{89.38$\pm$0.19} \\
		\bottomrule
	\end{tabular}}
	\label{tab:scratch}
\end{table*}

\subsubsection{Learning from the fixed teacher} \label{exp:fixed}

In this section, we fix the pretrained teacher network and only fine-tune the student network. This is the scheme used by~\cite{hinton2015distilling} to train their student network.
The training details for the student network are the same as those described in Sec.~\ref{exp:joint}.

From Table~\ref{tab:fixed_teacher}, we can observe that the improvement is relatively lower than that with jointly updated teachers in Table~\ref{tab:joint_PreResNet}. This proves that directly transferring the knowledge from the fixed pretrained teacher may not be optimal or not be suitable for quantization. Both $\Ml$ and $\Mf$ should be jointly optimized to adapt to each other. 

\begin{figure}[!htb]
	\centering
	\resizebox{0.8\linewidth}{!}
	{
		\begin{tabular}{c}
			\includegraphics{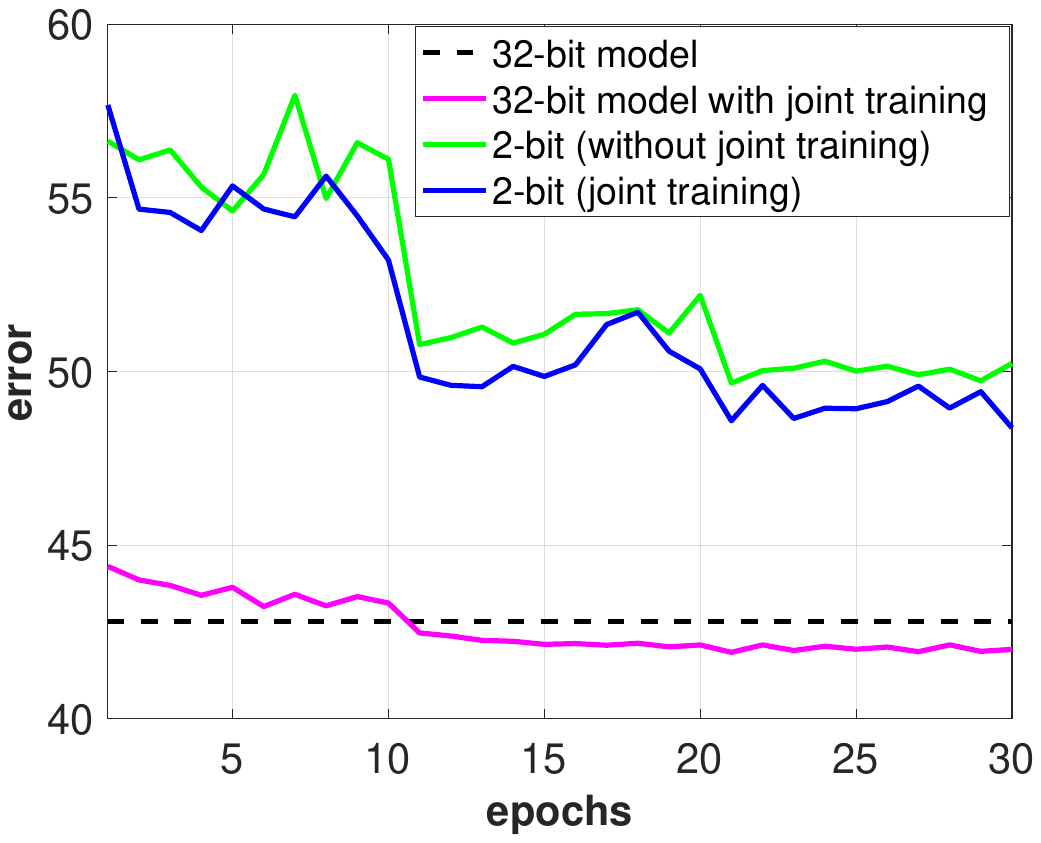}
		\end{tabular}
	}
	\caption{Joint training vs.\  fixed teacher using AlexNet* on ImageNet.}

	\label{fig:jointornot}
\end{figure}

However, this scheme has an advantage that one can pre-compute and store the guidance signals and access them during training $\Ml$, which can save the forward and backward pass computations w.r.t.\  $\Mf$.
For better understanding, we further show the convergence curves for AlexNet* on ImageNet in Figure~\ref{fig:jointornot}. 

\begin{table*}[!htb]
	\centering
	\small
	\caption{The accuracy of the quantized PreResNet using the fixed full-precision teacher.}
	\scalebox{0.9}
	{
	\begin{tabular}{cccccc}
		\toprule
		Precision & & \tabincell{c}{PreResNet-18 \\ Baseline} & \tabincell{c} {PreResNet-18 \\ with PreResNet-18} & \tabincell{c} {PreResNet-34 \\ Baseline}  & \tabincell{c} {PreResNet-34 \\ with PreResNet-34}  \\ \midrule
	    \multirow{2}{*}{4W, 4A} &Top-1\% &69.81  &\bf{70.15}  &73.57  &\bf{73.97}  \\ 
	     &Top-5\% &89.07  &\bf{89.48} &91.35 &\bf{91.72} \\ \midrule
	    \multirow{2}{*}{2W, 2A} &Top-1\% &64.51  &\bf{65.09}  &69.31  &\bf{69.96}  \\ 
	     &Top-5\% &85.85  &\bf{86.44}  &88.92  &\bf{89.51} \\
		\bottomrule
	\end{tabular}}
	\label{tab:fixed_teacher}
\end{table*}

\subsubsection{Ablation study on guidance signals}
\label{exp:hintvsposterior}
In this part, we further explore the effect of different distillation guidance signals as introduced in Sec.~\ref{sec:mutual}. The results are reported in Table~\ref{tab:hintvsposterior}. 
We observe that integrating both posterior-based and attention-based distillation strategies achieve the best result, performing better than using them separately.

\begin{table}[b]
	\centering
	\small
	\caption{Abalation study on the guidance signals with ResNet-50 on ImageNet.}
	\scalebox{0.75}
	{
	\begin{tabular}{cccccc}
		\toprule
		Precision & & \tabincell{c}{ResNet-50 \\ Baseline} & \tabincell{c} {ResNet-50 \\ posterior} & \tabincell{c} {ResNet-50  \\ attention transfer}  & \tabincell{c} {ResNet-50 \\ joint}  \\ \midrule
	    \multirow{2}{*}{2W, 2A} &Top-1\% &70.19 &71.40  &71.51   &\bf{71.96}   \\ 
	     &Top-5\% &89.15   &90.04   &90.17   &\bf{90.63}  \\
		\bottomrule
	\end{tabular}}
	\label{tab:hintvsposterior}
\end{table}

\subsubsection{Visualization}

\begin{figure}[t]
	\centering
	\resizebox{0.898\linewidth}{!}
	{
		\begin{tabular}{c}
			\includegraphics{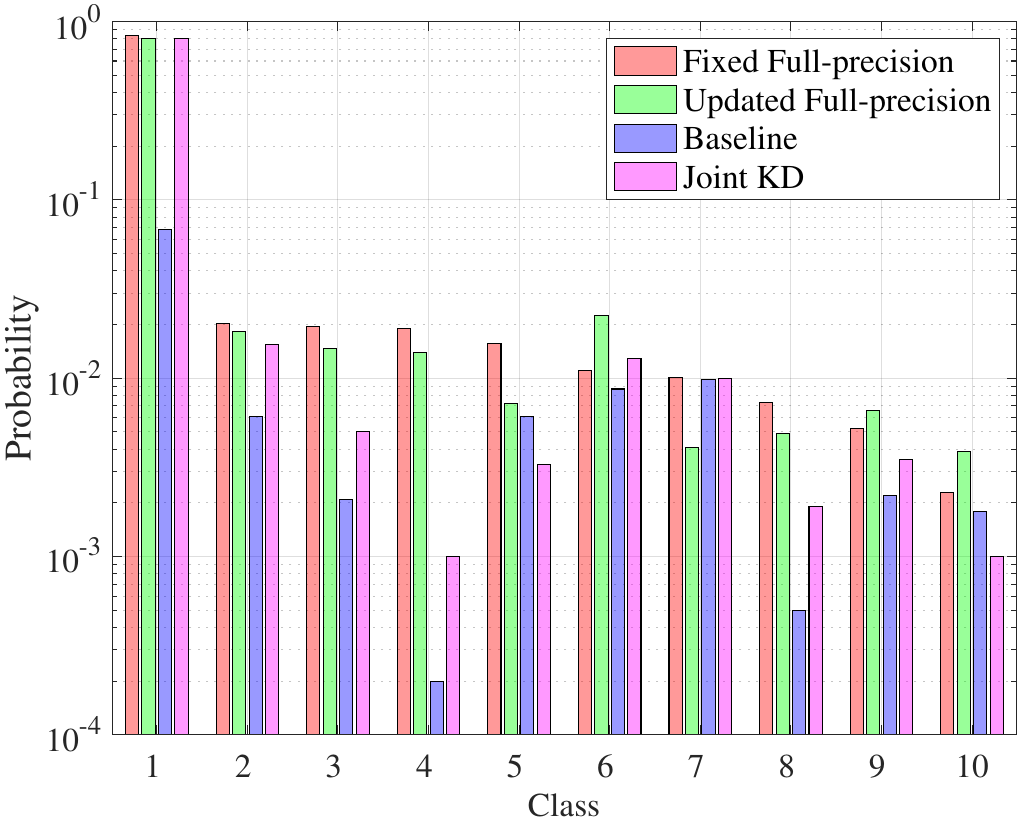}
		\end{tabular}
	}
	\caption{Mean posterior probability visualization.}
	\label{fig:visualization}
\end{figure}

We further visualize and conduct more analysis on the experimental results in Figure \ref{fig:visualization}. To explain why the proposed joint distillation strategy works better than the baseline, we illustrate the probability estimates assigned to Top-10 highest ranked classes obtained by a ResNet-18 on ImageNet trained by our joint distillation vs. an independently trained counterpart. This visualization is based on a randomly sampled mini-batch of 256. From Figure \ref{fig:visualization}, we have two main observations. First, we see that the posterior distribution of our proposed joint KD fits the full-precision counterpart better, which expects to have more accurate predictions. Second, the posterior probability of the jointly updated full-precision teacher adapts better to the low-precision student than the fixed full-precision teacher. This observation justifies that, with our joint distillation strategy, the low-precision student and the full-precision teacher learn collaboratively and adapt to each other throughout the training process. 

\subsection{Effect of quantizing all layers} \label{exp:all_layers}
In this part, we further explore the effect of quantizing the first convolution layer and the last classification layer to the final performance. We report the performance in Tables~\ref{tab:exp_twostep_progressive}, \ref{tab:whole_ResNet50_imagenet}, \ref{tab:whole_AlexNet_cifar} and 
\ref{tab:whole_AlexNet_imagenet}. With ``2W, 2A'', the performance of ResNet-50 beats ResNet-50* by a large margin. This shows that keeping the first and the last layer to high-precision is crucial to preserve the quantized model accuracy.
Moreover, the proposed advanced training approaches improve the baseline significantly. For 2-bit precision, the gap between ``ResNet-50* TS+PP+KD'' and baseline* is 2.93\% while ``ResNet-50 TS+PP+KD'' improves baseline by 2.21\%. It further justifies the claim in Sec.~\ref{exp:joint} and Sec.~\ref{exp:scratch} that the proposed training algorithms can be more effective when the model is more challenging to be optimized.

\subsection{Combining different training strategies}
\begin{table*}[!htb] 
	\centering
	\caption{Accuracy (\%) of ResNets on the ImageNet validation set using SP and KD.  Experiments are repeated for 3 times and we report the results with mean and standard deviation.}
	\scalebox{1.0}
	{
	\begin{tabular}{cccccccc}
		\toprule
		Precision & &\tabincell{c}{ResNet-18 \\ Baseline} & \tabincell{c} {ResNet-18 \\ with ResNet-18}  & \tabincell{c} {ResNet-34 \\ Baseline} &
		\tabincell{c} {ResNet-34 \\ with ResNet-34}  &\tabincell{c} {ResNet-50 \\ Baseline} & \tabincell{c} {ResNet-50 \\ with ResNet-50}  \\ \midrule
	    \multirow{2}{*}{32W, 32A} &Top-1\% &69.75  &- &73.21 &- &75.64 &-  \\ 
	     &Top-5\% &89.01  &- &91.40 &- &92.25 &- \\ \midrule
	    \multirow{2}{*}{2W, 2A} &Top-1\% &64.67$\pm$0.04 &\bf{66.07$\pm$}0.06  &68.17$\pm$0.05  &\bf{70.33$\pm$}0.05 &70.19$\pm$0.04 &\bf{73.25$\pm$}0.06 \\ 
	     &Top-5\% &85.78$\pm$0.10 &\bf{87.16$\pm$0.12} &88.05$\pm$0.09 &\bf{90.03$\pm$0.14} &89.15$\pm$0.08 &\bf{91.60$\pm$}0.15  \\
		\bottomrule
	\end{tabular}}
	\label{tab:whole_ResNet}
\end{table*}

\begin{table*}[!htb] 
	\centering
	\caption{Accuracy (\%) of PreResNets on the ImageNet validation set with SP and KD.  Experiments are repeated for 3 times and we report the results with mean and standard deviation.}
	\scalebox{1.0}
	{
	\begin{tabular}{cccccccc}
		\toprule
		Precision & & \tabincell{c}{PreResNet-18 \\ Baseline} & \tabincell{c} {PreResNet-18 \\ with ResNet-18}  & \tabincell{c} {PreResNet-34 \\ Baseline} &
		\tabincell{c} {PreResNet-34 \\ with PreResNet-34}  &\tabincell{c} {PreResNet-50 \\ Baseline} & \tabincell{c} {PreResNet-50 \\ with PreResNet-50}  \\ \midrule
	    \multirow{2}{*}{32W, 32A} &Top-1\% &69.95   &- &73.53  &- &76.11  &-  \\ 
	     &Top-5\% &89.21  &- &91.30  &- &92.81  &- \\ \midrule
	    \multirow{2}{*}{2W, 2A} &Top-1\% &64.51$\pm$0.05  &\bf{66.17$\pm$0.07}  &69.31$\pm$0.05  &\bf{70.65$\pm$0.06}  &71.20$\pm$0.04 &\bf{72.96$\pm$0.08} \\ 
	     &Top-5\% &85.85$\pm$0.10 &\bf{87.14$\pm$0.13}  &88.92$\pm$0.11 &\bf{89.72$\pm$0.16}  &90.18$\pm$0.10 &\bf{91.31$\pm$0.14}  \\
		\bottomrule
	\end{tabular}}
	\label{tab:whole_PreResNet}
\end{table*}

\begin{table}[!htb] 
	\centering
	\caption{Accuracy (\%) of ResNet-50 and ResNet-50* on the ImageNet with TS, PP and KD.}
	\scalebox{0.8}
	{
	\begin{tabular}{cccccc}
		\toprule
		Precision & & \tabincell{c}{Baseline*} &\tabincell{c} {ResNet-50* \\ TS + PP+ KD} &Baseline & \tabincell{c} {ResNet-50 \\ TS + PP + KD}  \\ \midrule
	    \multirow{2}{*}{32W, 32A} &Top-1\% &75.64  &- &- &- \\ 
	     &Top-5\% &92.25  &- &- &- \\ \midrule
	     \multirow{2}{*}{4W, 4A} &Top-1\% &74.30 &\bf{75.78} &74.50 &\bf{75.85} \\ 
	     &Top-5\% &91.16  &\bf{92.08} &91.46 &\bf{92.27} \\ \midrule
	    \multirow{2}{*}{2W, 2A} &Top-1\% &67.69  &\bf{70.62} &70.19  &\bf{72.40} \\ 
	     &Top-5\% &84.71  &\bf{88.07} &89.15 &\bf{90.65} \\
		\bottomrule
	\end{tabular}}
	\label{tab:whole_ResNet50_imagenet}
\end{table}

\begin{table}[!htb] 
	\centering
	\caption{Accuracy (\%) of AlexNet* on the CIFAR-100 with SP and KD.}
	\scalebox{1.0}
	{
	\begin{tabular}{cccc}
		\toprule
		Precision & & \tabincell{c}{AlexNet* \\ Baseline} & \tabincell{c} {AlexNet* \\ SP + KD}  \\ \midrule
	    \multirow{2}{*}{32W, 32A} &Top-1\% &65.42  &-  \\ 
	     &Top-5\% &88.31  &- \\ \midrule
	    \multirow{2}{*}{2W, 2A} &Top-1\% &63.89  &\bf{65.23}  \\ 
	     &Top-5\% &87.58 &\bf{88.44}  \\
		\bottomrule
	\end{tabular}}
	\label{tab:whole_AlexNet_cifar}
\end{table}

\begin{table}[!htb] 
	\centering
	\caption{Accuracy (\%) of AlexNet* on the ImageNet with TS, PP and KD.}
	\scalebox{1.0}
	{
	\begin{tabular}{ccccc}
		\toprule
		Precision & & \tabincell{c}{AlexNet* \\ Baseline} &\tabincell{c} {AlexNet* \\ TS + KD} & \tabincell{c} {AlexNet* \\ TS + PP + KD}  \\ \midrule
	    \multirow{2}{*}{32W, 32A} &Top-1\% &57.22  &- &-  \\ 
	     &Top-5\% &80.32  &- &- \\ \midrule
	     \multirow{2}{*}{4W, 4A} &Top-1\% &56.80  &57.92  &\bf{58.21} \\ 
	     &Top-5\% &80.01  &80.99  &\bf{81.28} \\ \midrule
	    \multirow{2}{*}{2W, 2A} &Top-1\% &48.79  &51.38  &\bf{51.96} \\ 
	     &Top-5\% &72.24  &75.60  &\bf{76.53} \\
		\bottomrule
	\end{tabular}}
	\label{tab:whole_AlexNet_imagenet}
\end{table}

Finally, we come to our complete approach by combining TS, PP, SP and KD. 
We first combine TS, PP with KD and the results are shown in Tables~\ref{tab:whole_ResNet50_imagenet} and \ref{tab:whole_AlexNet_imagenet}.
Moreover, we also combine the one-stage SP strategy with KD and the full results are reported in Tables~\ref{tab:whole_ResNet}, \ref{tab:whole_PreResNet} and \ref{tab:whole_AlexNet_cifar}.

We  observe that the proposed approaches can benefit from each other and further improve the performance on all settings. For instance, with ``2W, 2A" in Table~\ref{tab:whole_ResNet}, we find a 3.06\% relative gap between the baseline on ResNet-50. Even with the basic quantizer in DoReFa-Net, the difference in Top-1 error is only 2.39\%. 
This strongly justifies that the proposed joint knowledge distillation and the stochastic precision are general training approaches for improving low-bit neural networks.

\section{Conclusion}
In this paper, we have proposed three novel approaches to solve the optimization problem for quantizing the network with both low-precision weights and activations. 
Firstly, we have proposed the progressive quantization approach which includes two schemes.
Specifically, we have proposed a two-stage training scheme, where we use the real-valued activations as an intermediate step. We have also observed that continuously quantizing from high-precision to low-precision is also beneficial to the final performance. 
Moreover, we have proposed a stochastic precision strategy to significantly reduce the training complexity of progressive quantization while still improving the performance.

Furthermore, we have presented to improve the accuracy of low-precision networks with knowledge distillation. 
In particular, to better take advantage of the knowledge from the full-precision model, we have proposed to jointly learn the low-precision model and its full-precision counterpart. We have explored various distillation schemes and all observed improvements over the baseline.
Finally, we have combined the three training approaches to further boost the performance. We have conducted extensive experiments to justify the effectiveness of the proposed approaches on the image classification task.

\ifCLASSOPTIONcompsoc
  \section*{Acknowledgments}
\else
  \section*{Acknowledgment}
\fi

M. Tan was partially supported by National Natural Science Foundation of China (NSFC) 61602185, Program for Guangdong Introducing Innovative and Enterpreneurial Teams 2017ZT07X183.

\ifCLASSOPTIONcaptionsoff
  \newpage
\fi

\bibliographystyle{IEEEtran}
\bibliography{reference}

\begin{IEEEbiographynophoto}
{Bohan Zhuang}
is a Lecturer at Monash University, Australia. He received his PhD degree from The University of Adelaide.
\end{IEEEbiographynophoto}

\vspace{-.3cm}

\begin{IEEEbiographynophoto}
{Mingkui Tan}
is a Professor at South China University of Technology, China.
\end{IEEEbiographynophoto}
\vspace{-.3cm}

\begin{IEEEbiographynophoto}
{Jing Liu}
is 
a PhD student at Monash University.
\end{IEEEbiographynophoto}
\vspace{-.3cm}

\begin{IEEEbiographynophoto}
{Lingqiao Liu}
is a Senior Lecturer at The University of Adelaide.
\end{IEEEbiographynophoto}
\vspace{-.3cm}

\begin{IEEEbiographynophoto}
{Ian Reid}
is a Professor at The University of Adelaide.
\end{IEEEbiographynophoto}
\vspace{-.3cm}

\begin{IEEEbiographynophoto}
{Chunhua Shen}
%
is an Adjunct Professor
at Monash University.
\end{IEEEbiographynophoto}

\vfill

\end{document}